# AI/ML-Based Automatic Modulation Recognition: Recent Trends and Future Possibilities

Elaheh Jafarigol[1], Behnoud Alaghband[2], Azadeh Gilanpour[3], Saeid Hosseinipoor[4], and Mirhamed Mirmozafari[5]

[1]Data Science and Analytics Institute, University of Oklahoma, 202 W. Boyd St., Room 409, Norman, 73019, Ok, USA
[2]Electrical and Computer Science Department, University of New Mexico, Farris Engineering Center 1901 Redondo,MSC01 1130,Albuquerque, NM 87131,USA
[3]Labcorp Drug Development, 3301 Kinsman Blvd, Madison, WI 53704
[4]Samsung Research America, San Francisco, CA 94111
[5]Maxwave LLC, 2616 Dublin Way, Waunakee, WI, 53597

Corresponding author: Mirhamed Mirmozafari: mirmozafari@maxwllc.com).

**ABSTRACT** We present a review of high-performance automatic modulation recognition (AMR) models proposed in the literature to classify various Radio Frequency (RF) modulation schemes. We replicated these models and compared their performance in terms of accuracy across a range of signal-to-noise ratios. To ensure a fair comparison, we used the same dataset (RadioML-2016A), the same hardware, and a consistent definition of test accuracy as the evaluation metric, thereby providing a benchmark for future AMR studies. The hyperparameters were selected based on the authors' suggestions in the associated references to achieve results as close as possible to the originals. The replicated models are publicly accessible for further analysis of AMR models. We also present the test accuracies of the selected models versus their number of parameters, indicating their complexities. Building on this comparative analysis, we identify strategies to enhance these models' performance. Finally, we present potential opportunities for improvement, whether through novel architectures, data processing techniques, or training strategies, to further advance the capabilities of AMR models.

## I. INTRODUCTION

Automatic modulation recognition (AMR) serves as a critical component in modern communication systems, enabling the identification and classification of various modulation schemes in radio frequency (RF) signals without human intervention [1]. AMR plays a crucial role in spectrum monitoring as it is becoming increasingly congested due to the proliferation of wireless technologies. Especially, in the form of cognitive radio, AMR enables a rapid sweep of the frequency spectrum and identification of existing RF signals transmitted from friend or foe [2], [3]. By swiftly discerning modulation types, AMR facilitates spectrum utilization optimization and interference mitigation, thereby enhancing the efficiency, r eliability, a nd s ecurity of communication networks [4]. Its significance extends across diverse domains, including military communications, civilian telecommunications, wireless sensor networks, and the Internet of Things (IoT), underscoring its essential role in ensuring seamless and robust wireless connectivity in the modern era [5].

While traditional methods have their merits and have been effectively used in AMR for years, artificial intelligence and machine learning (AI/ML) models offer significant advantages in terms of accuracy, robustness, and adaptability [6], [7]. The automated feature extraction and ability to learn complex patterns make AI models a powerful tool for modern AMR problems, leading to better performance in real-world applications [8]. These intricate features cannot be mathematically formulated in a straightforward manner, thereby their rapid extraction enables real-time classification of RF signals with enhanced accuracy. Additionally, these AI/ML-based models can concurrently handle a wide range of modulation schemes, contributing to the scalability of such systems [9], [10]. Finally, the rapid development of computational hardware, along with their cost reduction, will soon facilitate the adoption of more complex AMR models in both research and field experiments.

The adoption of AI/ML-based models in AMR is not without challenges. These models rely heavily on large-scale labeled datasets for effective training. Therefore, a judiciously prepared dataset would considerably increase the robustness of the models. The dataset should include sufficient information for effective training of the AI/ML models while representing real-world scenarios for enhanced and practical performance. Moreover, the computational complexity associated with both training and deploying such models poses additional hurdles. Yet, perhaps the most critical challenge arises from the compromised performance of AI/ML-based models in low signal-to-noise ratio (SNR) scenarios, similar to non-AI/ML models [11]. In such situations, where the RF signal is weak or obscured by noise, the models struggle to deliver accurate results, highlighting the need for robust solutions. Furthermore, challenges related to execution time and the consistent performance of these models across diverse operational scenarios further contribute to the complexity [12]. Addressing these challenges is imperative for unlocking the full potential of AI/ML-based models.

This paper provides a comprehensive review of state-of-the-art AI/ML-based AMR models, with a focus on identifying innovative approaches for further development. Through meticulous comparative analysis, we assess the performance metrics of top-tier AMR models, maintaining a consistent standard for fair comparison. Furthermore, we will discuss modifications to a model to enhance its performance, which can be applied to others. This is aimed at determining the enhanced capabilities of recently proposed models by identifying key challenges that constrain their performance. Finally, by exploring avenues for performance enhancement, we propose novel approaches and strategies tailored to address the identified limitations, thus opening up new opportunities for future research and development in AMR field.

## II. AMR AI/ML MODELS AND COMPARATIVE ANALYSIS

Here, we selected several high-performance models from the literature representing state-of-the-art AI/ML-based models [13], as listed in Table 1. We replicated all these models and conducted a comparative analysis of their performance. We made the replicated models publicly available for easy reproducibility and adjustment, aiding further research and development of AMR modules. We evaluated the performance of these models under consistent conditions. These conditions were maintained throughout the execution of the models and the reporting of results, ensuring a fair and accurate comparative analysis. Specifically, we used the same dataset and identical data split percentages for training, validation, and test. More importantly, we adhered to a uniform evaluation criterion, to guarantee that comparisons between the different models were fair and consistent.

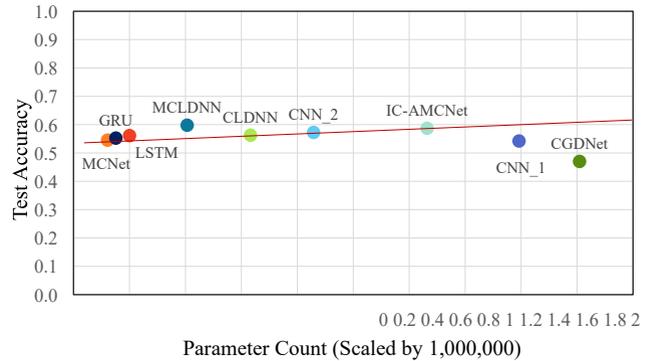

FIGURE 1: The comparison of models.

The RadioML-2016A dataset [14] is used to benchmark the models. The synthetic nature of the dataset ensures reproducibility and scalability, making it a standard source for training and evaluating AI/ML-based models discussed below. This dataset consists of synthetic IQ (In-phase and Quadrature) samples representing various modulation schemes, including amplitude modulation (AM), frequency modulation (FM), and phase modulation (PM) across different SNR levels [15]. 11 modulation types, uniformly distributed across SNR levels, ranging from -20 dB to 18 dB are incorporated in the RadioML-2016 dataset. The data is split 60%, 20%, and 20% for training, validation, and testing, respectively. The hyper-parameters such as learning rate and batch size are selected based on the authors' suggestions in the associated references. The models are run on GPU on Google Colab. The models are trained up to 100 epochs with early stopping on validation loss. After executing the models, we calculated the accuracy, which represents the percentage of predicted values that match the actual values for one-hot-encoded labels. In this document, we report the test accuracy for each model as a uniform evaluation criterion, derived from average accuracy values across all modulation schemes and all SNR levels. This consistent comparison criterion ensures fair evaluations of the models by avoiding selective reporting of this parameter, thereby providing clear and unbiased results. Detailed execution and evaluation parameters are provided in Table 1.

Figure 1 illustrates the test accuracy values for selected models versus their number of trainable parameters. It is commonly expected that higher parameter counts lead to better performance, as indicated by the fitted red line in Figure 1, which serves as a threshold to compare the performance of models with various sizes. However, the data suggest a slightly different trend [25]. MCLDNN and IC-AMCNet stand out for achieving high accuracy with a moderate parameter count. This performance is valuable in applications where virtually real-time recognition is needed while computational resources are limited. The findings emphasize the importance of optimizing model architecture for AMR applications. For instance, CLDNN and LSTM demonstrate very good performance with relatively few parameters, suggesting that incorporating recurrent layers or combining convolutional and recurrent layers can enhance model effectiveness in AMR tasks. This highlights practical approaches to balancing complexity and performance, particularly in resource-constrained environments [26], [27].

Figure 2 addresses the issue of noise in AMR models, which becomes a significant challenge as its level increases relative to signal values. The presence of noise notably impacts the accuracy of AMR models, with performance generally deteriorating as the SNR decreases. Figure 2 displays the test accuracy of the models versus SNR ratios for all modulation classes. We see three distinct zones, where all models almost show similar behavior. At very low SNR levels (from -20 dB to -12 dB) [28], the test accuracy of all models is significantly low, with performance below 0.3. This indicates that noise significantly impairs the models' 

TABLE 1: Execution and evaluation parameters for our comparative analysis of selected AMR models.

| Model | | Parameter Count | Batch size | Learning rate | Epochs | Test Accuracy |
|---|---|---|---|---|---|---|
| CNN1 | [16] | 1,592,383 | 1024 | 1e-4 | 100 | 0.5423 |
| CNN2 | [17] | 858,123 | 1024 | 1e-4 | 100 | 0.5725 |
| CLDNN | [18] | 632,531 | 400 | 1e-3 | 100 | 0.5650 |
| IC-AMCNet | [19] | 1,264,011 | 400 | 1e-3 | 77 | 0.5870 |
| MCNet | [20] | 121,611 | 128 | 1e-4 | 63 | 0.5453 |
| LSTM | [21] | 200,075 | 400 | 1e-3 | 84 | 0.5615 |
| GRU | [22] | 151,179 | 400 | 1e-3 | 54 | 0.5526 |
| MCLDNN | [23] | 405,887 | 400 | 1e-3 | 39 | 0.5982 |
| CGDNet | [24] | 1,808,811 | 1024 | 1e-2 | 42 | 0.4700 |

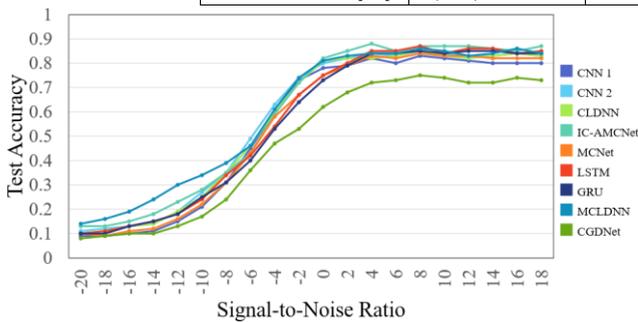

FIGURE 2: Test accuracy of replicated AMR models across various SNR levels, with each SNR value encompassing all modulation schemes.

simpler models can perform well. Overall, MCLDNN and IC-AMCNet show greater resilience and higher accuracy in low to moderate SNR ranges compared to other models. This suggests that these architectures are more robust and better suited for handling noisy conditions.

The analysis reveals that noisy environments can mask the intrinsic properties of the modulation signal that AMR models rely on for correct classification. This is particularly problematic in low-SNR scenarios where the noise levels are high relative to the signal, making it difficult for ability to classify signals accurately. As the SNR increases from -12 dB to 0 dB, there is a gradual improvement in the test accuracy [29]. This movement suggests that the models begin to extract meaningful features from the signals as the noise decreases, however, the classification r emains challenging. Accuracies rise significantly at moderate SNR levels between 0 dB and 6 dB, where the models leverage the improving signal quality to enhance their classification performance, and the curves start to plateau as the SNR increases beyond 6. The convergence of performance across different models at higher SNR levels suggests that when the signal is clear, even the model to differentiate between noise-induced variations and genuine modulation characteristics. This limitation necessitates robust feature extraction and enhanced learning algorithms that are resilient to noise and capable of extracting and leveraging subtle differences in the signal, even under adverse conditions.

Table 2 demonstrates the test accuracy of the models across different modulation types.

The analysis of test accuracy across 11 modulation types reveals several important trends. First, certain modulation schemes are generally difficult to recognize using current AMR models. For instance, most models struggle to recognize GFSK, with the LSTM model achieving the best accuracy at just 0.42. On the other hand, QPSK is relatively easy to recognize, with most models yielding an accuracy of over 0.9. This highlights the need to develop specialized models focused on enhancing recognition accuracy for these hard-to-detect modulations. The second observation is that different models excel in various areas, but no single model dominates across all modulation schemes. For instance, CNN 1 and CLDNN perform exceptionally well on QPSK, both achieving a perfect accuracy of 0.99, indicating their strong capability in handling phase modulation. However, these models struggle significantly with GFSK, where CLDNN even fails to classify correctly (close to 0.00 accuracy), highlighting the challenges posed by frequency-shift keying modulations. This variability in model performance suggests that different architectures have strengths and weaknesses depending on the modulation type. Therefore, hybrid approaches also referred to as ensemble learning, can combine the strengths of multiple models and could potentially be beneficial in ensuring comprehensive and reliable performance across all modulation types [30].

data preparation for training and testing or through modifications to the model structure.

*A. INVESTIGATING TRAINING APPROACHES ON A NOISY DATASET*

Here, we introduce an approach to partially address the noise issue in modulation recognition. Specifically, we present a training method that enables models to learn more intricate features of modulated signals, thereby mitigating the impact of noise. For this study, MCLDNN which offers the best performance among other models was selected to check if the ultimate performance of AMR models can be improved. In this approach, we do not train the model on the entire dataset, which includes all SNR values ranging from -20 to 18 dB. Instead, we preserved the high SNR data (0 to 18 dB) and gradually added low SNR data from -2 to -20 dB. Once the model was trained, testing was conducted across all SNR values.

Figure 3 summarizes the performance of the MCLDNN model trained and tested under various SNR conditions [31]. The horizontal axis represents different training scenarios, with each case showing the range of SNR values used for training purposes. Three vertical bars are presented for each training scenario. From left to right, these bars represent the test accuracy for low SNR data (SNR < 0), the test accuracy for high SNR data (SNR > 0), and the average test accuracy for the entire test dataset [32]. The test reveals several informative trends. Models trained on high SNR data achieve the highest test accuracy when evaluated on high SNR data, suggesting that clean, high-quality signals provide the most useful information for training robust models. Models trained on a mixture of SNR levels achieve the best overall generalization, performing moderately well across both low and high SNR conditions. This balance indicates the importance of exposing the model to a variety of noise levels during training. The most important conclusion is that the highest accuracy (0.63) is achieved with training data ranging from -18 to 18 dB, rather than using the entire range of SNR

TABLE 2: Test accuracy of replicated AMR models for different modulation schemes. Each test accuracy value encompasses all SNR values of that specific modulation.

| Model | WBFM | AM-DSB | BPSK | QPSK | 8PSK | QAM64 | CPFSK | AM-SSB | PAM4 | GFSK | QAM16 |
|---|---|---|---|---|---|---|---|---|---|---|---|
| CNN 1 | 0.40 | 0.70 | 0.64 | 0.99 | 0.46 | 0.51 | 0.61 | 0.69 | 0.67 | 0.04 | 0.57 |
| CNN 2 | 0.42 | 0.49 | 0.67 | 0.96 | 0.55 | 0.58 | 0.61 | 0.65 | 0.64 | 0.23 | 0.57 |
| CLDNN | 0.67 | 0.40 | 0.63 | 0.99 | 0.55 | 0.52 | 0.62 | 0.68 | 0.63 | 0.00 | 0.52 |
| IC-AMCNet | 0.30 | 0.79 | 0.62 | 0.97 | 0.54 | 0.33 | 0.64 | 0.68 | 0.68 | 0.35 | 0.56 |
| MCNet | 0.35 | 0.73 | 0.65 | 0.89 | 0.43 | 0.27 | 0.59 | 0.65 | 0.61 | 0.34 | 0.47 |
| LSTM | 0.39 | 0.71 | 0.61 | 0.91 | 0.49 | 0.27 | 0.57 | 0.67 | 0.62 | 0.42 | 0.50 |
| GRU | 0.39 | 0.70 | 0.59 | 0.90 | 0.51 | 0.30 | 0.57 | 0.67 | 0.63 | 0.36 | 0.44 |
| MCLDNN | 0.47 | 0.75 | 0.65 | 0.93 | 0.52 | 0.58 | 0.62 | 0.67 | 0.72 | 0.03 | 0.64 |
| CGDNet | 0.26 | 0.75 | 0.57 | 0.95 | 0.21 | 0.39 | 0.52 | 0.55 | 0.50 | 0.06 | 0.39 |

## III. PERFORMANCE ENHANCEMENT OF AI/ML-BASED MODELS

Building on discoveries found in Section I, we present adjustments to some of the above-mentioned models that improve their performance. These adjustments are either in values. This suggests that noise-heavy training data limits the model's ability to learn useful signal features, impacting its performance on cleaner data.

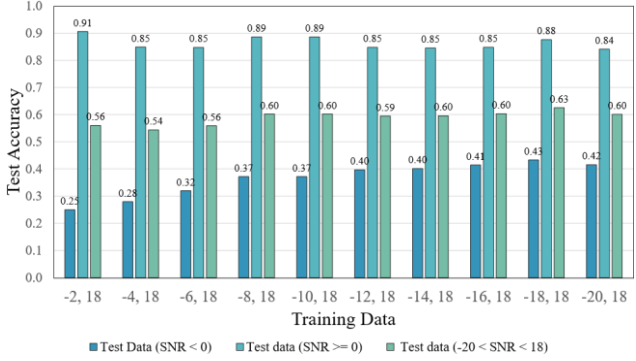

FIGURE 3: MCLDNN is trained and tested under high, low, and full range of SNR levels

Therefore, we observe the significance of tailored training approaches for optimum performance in noisy environments. This suggests that the optimal training approach requires a careful selection of training datasets to maximize each model's ability to recognize various modulation characteristics. This may involve expanding beyond the current dataset, and these opportunities will be discussed in Section III.

### B. ARCHITECTURE VARIATIONS

Here, we explore the enhancement of model architectures through the addition of extra layers to the original models. The original models are those discussed in Section II and additional layers were designed to improve feature extraction capabilities. These additional layers serve to complement and augment the original models, ultimately enhancing their overall performance. We chose to add Gated Recurrent Units (GRUs) and Bidirectional LSTM (BiLSTM) layers, whose specific characteristics serve this purpose [33]. These models, which are relatively underexplored in this context, are known for their longer memory capacities, enabling the system to effectively retain and leverage historical data. By analyzing sequences of data over extended periods, these layers can identify correlations between data points within a group, thereby enhancing the model's ability to recall and utilize historical information. This improved memory and correlation capability significantly boosts the model's performance in predicting new modulations, leading to a more robust and accurate overall performance [34].

GRU [35], a variant of Recurrent Neural Networks (RNNs), which is a simplified version of LSTM efficiently captures temporal dependencies in sequential data, making them well-suited for processing time-series data like IQ samples. When combined with CNN layers, GRUs leverage spatial feature extraction alongside temporal pattern recognition, resulting in a more comprehensive understanding of signal characteristics [36].

BiLSTM layers, which process input data in both forward and backward directions, allow the model to access future context, further improving its ability to detect and classify modulation patterns accurately. This approach ensures that the network fully utilizes the information available in the entire sequence, enhancing the robustness and generalization of the model across diverse SNR conditions [37].

We added BiLSTM and GRUs to all models listed in Table 1 and executed the modified models. To ensure a fair comparison, we ran the modified models with the same batch size and learning rates as listed in Table 1. The test accuracy results for the modified models are presented in Table 3, alongside the original test accuracy results from the last column of Table 1.

We observed an enhancement in the performance of some models, specifically CLDNN, IC-AMCNet, and MCLDNN. Adding BiLSTM and GRUs layers appears to be an effective complement for these models, enhancing their performance. Conversely, we saw no or negligible improvements for models like LSTM, CNN2, and MCNet [38], with slight differences in test accuracies attributed to the randomness of the procedure. This is partly due to fact that either some of these models like LSTM has the long-short memory features [39] already embedded into them.

TABLE 3: Comparison of test accuracy before and after adding BiLSTM and GRU Layers.

| Model | | Test Accuracy | Modified Test Accuracy |
|---|---|---|---|
| CNN1 | [16] | 0.5423 | 0.5523 |
| CNN2 | [17] | 0.5725 | 0.5633 |
| CLDNN | [18] | 0.5650 | 0.6158 |
| IC-AMCNet | [19] | 0.5870 | 0.6237 |
| MCNet | [20] | 0.5453 | 0.5432 |
| LSTM | [21] | 0.5615 | 0.5715 |
| GRU | [22] | 0.5526 | 0.5753 |
| MCLDNN | [23] | 0.5982 | 0.6580 |
| CGDNet | [24] | 0.4700 | 0.5043 |

## IV. FUTURE APPROACHES TO ELEVATE AMR PERFORMANCE

This section briefly discusses the future possibilities for enhancing the performance of AMR models. Building on the discoveries outlined in Section III, potential approaches include developing new architectures with improved feature identification capabilities, generating advanced supplementary datasets, and improving training strategies. These approaches can be specifically tailored to tackle noise issues in AMR models.

### A. DATA PREPROCESSING AND SYNTHETIC DATASET GENERATION

The information embedded in a dataset is crucial for training AMR models, and maximizing the extraction of this information can significantly improve model training [40]. This can be achieved through additional preprocessing techniques. For example, in [41], frequency-domain features were utilized alongside time-domain features to enhance model performance. Additionally, denoising modules can be incorporated into the preprocessing stage to filter out noise embedded in the test dataset. Furthermore, more sophisticated denoising modules, operating in either the time or frequency domain, can be implemented to further reduce noise effects, ultimately leading to more accurate recognition [42].

In addition to the available dataset, a wealth of information can be incorporated into a supplementary dataset. This supplementary dataset can capture the intricate features of various modulation schemes, significantly enhancing the training capability of AMR models. By feeding this synthetic dataset into the model alongside the original test dataset, complements the training process, enriching the model's ability to recognize and generalize features effectively.

By employing synthetic data generation methods [43], [44] we can design a curated dataset that could be employed for specific reasons. For example, we can add various noises with different distributions and characteristics and then design AI/ML models such as Generative Adversarial Networks (GANs) [45], Variational Autoencoders (VAEs) [46], to damp the noise of input signals. Therefore, the synthetic dataset can include a wide range of intentionally generated noisy features, which help to better train the model and increase its robustness in noisy environments.

Additional features in the supplementary dataset may include classic characteristics that facilitate faster model convergence by pre-training on the synthetic dataset and then fine-tuning on the target dataset which may also lead to more accurate classification [47], [48]. Furthermore, these supplementary datasets can be designed with non-uniform distributions, biased towards the more challenging modulation schemes. This targeted approach ensures that models are better equipped to handle difficult cases, ultimately leading to more robust and reliable modulation recognition in real-world applications.

### B. NEW ARCHITECTURE WITH ENHANCED FEATURES

Figure 2 highlights several limitations of previously proposed models. Not only do these models perform poorly in noisy environments, but they also fail to achieve 100% accuracy even with clean data. The plateau observed around 90% efficiency in the plots underscores this limitation. These shortcomings demonstrate the need for significantly improved architectures for advanced AMR.

Transformer is a new emerging model that proved its ability in sequence problems such as Large Language Models (LLM). Transformers have potential capabilities that make them suitable for AMR applications [49]. The possibility of incorporating attention mechanisms in transformers seems to be a great option for various modulation recognition. Transformers are capable of dynamically focusing on different parts of the input data, assigning varying levels of importance to each part based on the context. This selective attention to relevant features makes transformers particularly effective in handling complex and diverse modulation patterns [50].

Table 3 demonstrates a representative scenario where modifications to some existing architectures yield enhanced performance in accuracy. These enhancements are attributed to incorporating models that can memorize longer terms, allowing the system to remember and leverage data history. This feature enables the model to connect dots between further but more related data in the data sequence, leading to overall better performance. We demonstrated an illustrative example where we used models equipped with both long-term and short-term memory mechanisms [51]. Long Short-Term Memory (LSTM) networks, for example, are well-suited for capturing long-term dependencies, while Gated Recurrent Units (GRUs) offer a more streamlined approach to managing short-term dependencies. By retaining pertinent information over varying time spans, these models can provide a more nuanced understanding of the data, enhancing prediction accuracy. In our illustrative scenarios, these models are used as additional layers to augment existing architectures; however, they can also be deployed as standalone models with optimized hyperparameters and configurations.

## C. ENHANCED TRAINING STRATEGIES

Traditionally, AMR models are trained using a single test dataset sampled from an available data source. However, the creation of supplementary datasets, as discussed in IV-A, calls for more sophisticated training approaches that can effectively utilize multiple distributed datasets. Federated learning offers a distributed solution to this challenge by leveraging decentralized data sources [52]. Unlike traditional centralized training methods, federated learning allows multiple devices or nodes to collaboratively train a shared model while keeping their data locally. Each participating device updates the model using its local data, and only the model parameters are aggregated centrally. This process results in a global model that benefits from the diversity of decentralized data without compromising privacy.

Moreover, federated learning facilitates continuous learning, enabling AMR models to adapt dynamically to new signal environments and modulation schemes as they are encountered by different devices. This adaptability is crucial for maintaining high recognition accuracy amidst evolving communication technologies and diverse deployment scenarios. By harnessing the power of federated learning, AMR systems can achieve improved performance, enhanced privacy, and greater scalability, making them more robust and versatile for real-world applications [53].

Another promising training approach to enhance AMR performance is Ensemble Learning. This method is particularly effective in addressing the common challenge faced by many proposed models—the difficulty in accurately recognizing specific modulation schemes. Ensemble Learning can be implemented in several ways, each leveraging the strengths of individual models to improve overall performance. One approach involves creating multiple instances of the same architecture, each specialized for specific classes. These specialized models can then be combined through a voting mechanism, where the final prediction is determined by aggregating the outputs of the ensemble members. This approach enhances the model's robustness by reducing the likelihood of misclassification for challenging modulation schemes.

Alternatively, Ensemble Learning can be applied by selecting diverse models, each excelling at recognizing particular classes of modulations. These models are then strategically combined to maximize the overall classification performance. By harnessing the unique strengths of different models, this method addresses the inherent weaknesses of individual models, leading to more accurate and reliable AMR systems.

## V. CONCLUSION

In this study, we reviewed and replicated previously proposed AMR models, providing a comparative analysis of their performance under rigorously controlled conditions. Our evaluation, using a consistent data set and identical testing environments, revealed the strengths and limitations of these models. Notably, all models demonstrated weaknesses at low SNR values and exhibited capped performance even under high SNR conditions, indicating persistent recognition errors that prevent achieving 100% accuracy. Further investigation highlighted specific challenges with certain modulation schemes, which motivated us to explore targeted modifications.

Our modifications, including adjustments in training approaches and architectural refinements, led to enhanced performance, suggesting potential for further improvements. Finally, we present future avenues for further research and exploration in AMR, categorized into three key areas: data preprocessing, next-generation architectures, and advanced training methodologies. Each of these areas holds significant potential to enhance AMR performance, offering new opportunities for research and development.

...

**REFERENCES**

[1] Dong Yi, Di Wu, and Tao Hu. A lightweight automatic modulation recognition algorithm based on deep learning. IEICE Transactions on Communications, E106.B(4):367–373, April 2023.

[2] Fuxin Zhang, Chunbo Luo, Jialang Xu, Yang Luo, and Fu-Chun Zheng. Deep learning based automatic modulation recognition: Models, datasets, and challenges. Digital Signal Processing, 129:103650, September 2022.

[3] Kuiyu Chen, Si Chen, Shuning Zhang, and Huichang Zhao. Automatic modulation recognition of radar signals based on histogram of oriented gradient via improved principal component analysis. Signal Image and Video Processing, 17(6):3053–3061, April 2023.

[4] Xueyang Hu et al. Smart sensing and context-cognitive networking in and beyond mmwave band: Efficiency, reliability, and security. 2024.

[5] Seung-Hwan Kim, Jae-Woo Kim, Van-Sang Doan, and Dong-Seong Kim. Lightweight deep learning model for automatic modulation classification in cognitive radio networks. IEEE Access, 8:197532–197541, 2020.

[6] I. Evelyn Ezhilarasi, J. Christopher Clement, and Joseph M. Arul. A survey on cognitive radio network attack mitigation using machine learning and blockchain. EURASIP Journal on Wireless Communications and Networking, 2023(1), September 2023.

[7] Tara N Sainath, Ron J Weiss, Andrew W Senior, Kevin W Wilson, and Oriol Vinyals. Learning the speech front-end with raw waveform cldnns. In Interspeech, pages 1–5. Dresden, Germany, 2015.

[8] Yunpeng Qu, Zhilin Lu, Rui Zeng, Jintao Wang, and Jian Wang. Enhancing automatic modulation recognition through robust global feature extraction. arXiv preprint arXiv:2401.01056, 2024.

[9] Hang Zhang, Ruihua Nie, Minghui Lin, Ruijuan Wu, Guo Xian, Xiaofeng Gong, Qin Yu, and Ruisen Luo. A deep learning based algorithm with multi-level feature extraction for automatic modulation recognition. Wireless Networks, 27(7):4665–4676, 2021.

[10] Hilal Elyousseph and Majid L Altamimi. Deep learning radio frequency signal classification with hybrid images. In 2021 IEEE International Conference on Signal and Image Processing Applications (ICSIPA), pages 7–11. IEEE, 2021.

[11] Jiawei Zhang, Tiantian Wang, Zhixi Feng, and Shuyuan Yang. Amc-net: An effective network for automatic modulation classification. In ICASSP 2023-2023 IEEE International Conference on Acoustics, Speech and Signal Processing (ICASSP), pages 1–5. IEEE, 2023.

[12] Fuxin Zhang, Chunbo Luo, Jialang Xu, Yang Luo, and Fu-Chun Zheng. Deep learning based automatic modulation recognition: Models, datasets, and challenges. Digital Signal Processing, 129:103650, 2022.

[13] Ruolin Zhou, Fugang Liu, and Christopher W. Gravelle. Deep learning for modulation recognition: A survey with a demonstration. IEEE Access, 8:67366–67376, January 2020.

[14] Timothy James O'Shea, Tamoghna Roy, and T Charles Clancy. Over-the-air deep learning based radio signal classification. IEEE Journal of Selected Topics in Signal Processing, 12(1):168–179, 2018.

[15] Amir Shachar. Introduction to algogens, 2024.

[16] Timothy J O'Shea, Johnathan Corgan, and T Charles Clancy. Convolutional radio modulation recognition networks. In Engineering Applications of Neural Networks: 17th International Conference, EANN 2016, Aberdeen, UK, September 2-5, 2016, Proceedings 17, pages 213–226. Springer, 2016.

[17] Kürşat Tekbıyık, Ali Rıza Ekti, Ali Görçin, Güneş Karabulut Kurt, and Cihat Keçeci. Robust and fast automatic modulation classification with cnn under multipath fading channels. In 2020 IEEE 91st Vehicular Technology Conference (VTC2020-Spring), pages 1–6. IEEE, 2020.

[18] Nathan E West and Tim O'shea. Deep architectures for modulation recognition. In 2017 IEEE international symposium on dynamic spectrum access networks (DySPAN), pages 1–6. IEEE, 2017.

[19] Ade Pitra Hermawan, Rizki Rivai Ginanjar, Dong-Seong Kim, and Jae-Min Lee. Cnn-based automatic modulation classification for beyond 5g communications. IEEE Communications Letters, 24(5):1038–1041, 2020.

[20] Thien Huynh-The, Cam-Hao Hua, Quoc-Viet Pham, and Dong-Seong Kim. Mcnet: An efficient cnn architecture for robust automatic modulation classification. IEEE Communications Letters, 24(4):811–815, 2020.

[21] Sreeraj Rajendran, Wannes Meert, Domenico Giustiniano, Vincent Lenders, and Sofie Pollin. Deep learning models for wireless signal classification with distributed low-cost spectrum sensors. IEEE Transactions on Cognitive Communications and Networking, 4(3):433–445, 2018.

[22] Dehua Hong, Zilong Zhang, and Xiaodong Xu. Automatic modulation classification using recurrent neural networks. In 2017 3rd IEEE International Conference on Computer and Communications (ICCC), pages 695–700. IEEE, 2017.

[23] Jialang Xu, Chunbo Luo, Gerard Parr, and Yang Luo. A spatiotemporal multi-channel learning framework for automatic modulation recognition. IEEE Wireless Communications Letters, 9(10):1629–1632, 2020.

[24] Judith Nkechinyere Njoku, Manuel Eugenio Morocho-Cayamcela, and Wansu Lim. Cgdnet: Efficient hybrid deep learning model for robust automatic modulation recognition. IEEE Networking Letters, 3(2):47–51, 2021.

[25] Zhe Jiang, Jingbo Zhang, Tianxing Wang, and Haiyan Wang. Modulation recognition of underwater acoustic communication signals based on neural architecture search. Applied Acoustics, 225:110155, 2024.

[26] Fuxin Zhang, Chunbo Luo, Jialang Xu, and Yang Luo. An efficient deep learning model for automatic modulation recognition based on parameter estimation and transformation. IEEE Communications Letters, 25(10):3287–3290, 2021.

[27] Meng Qi, Nianfeng Shi, Guoqiang Wang, and Hongxiang Shao. Data-transform multi-channel hybrid deep learning for automatic modulation recognition. IEEE Access, 2024.

[28] None Mohammad Chegini, None Pouya Shiri, and None Amirali Baniasadi. Rfnet: Fast and efficient neural network for modulation classification of radio frequency signals. ITU Journal on Future and Evolving Technologies, 3(2):261–272, September 2022.

[29] Seung-Hwan Kim, Jae-Woo Kim, Van-Sang Doan, and Dong-Seong Kim. Lightweight deep learning model for automatic modulation classification in cognitive radio networks. IEEE Access, 8:197532–197541, January 2020.

[30] Ayman Emam, M Shalaby, Mohamed Atta Aboelazm, Hossam E Abou Bakr, and Hany AA Mansour. A comparative study between cnn, lstm, and cldnn models in the context of radio modulation classification. In 2020 12th International Conference on Electrical Engineering (ICEENG), pages 190–195. IEEE, 2020.

[31] Adeeb Abbas, Vasil Pano, Geoffrey Mainland, and Kapil Dandekar. Radio modulation classification using deep residual neural networks. In MILCOM 2022-2022 IEEE Military Communications Conference (MILCOM), pages 311–317. IEEE, 2022.

[32] Aykut Büker and Cemal Hanilçi. Deep convolutional neural networks for double compressed amr audio detection. IET Signal Processing, 15(4):265–280, 2021.

[33] Md Shofiqul Islam and Ngahzaifa Ab Ghani. A novel bigrubilstm model for multilevel sentiment analysis using deep neural network with bigru-bilstm. In Recent Trends in Mechatronics Towards Industry 4.0: Selected Articles from iM3F 2020, Malaysia, pages 403–414. Springer, 2021.

[34] Minsik Cho and Daniel Brand. Mec: Memory-efficient convolution for deep neural network. In International Conference on Machine Learning, pages 815–824. PMLR, 2017.

[35] Kyunghyun Cho, Bart van Merrienboer, Caglar Gulcehre, Dzmitry Bahdanau, Fethi Bougares, Holger Schwenk, and Yoshua Bengio. Learning phrase representations using rnn encoder-decoder for statistical machine translation, 2014.

[36] Bo Cao, Chenghai Li, Yafei Song, Yueyi Qin, and Chen Chen. Network intrusion detection model based on cnn and gru. Applied Sciences, 12(9):4184, 2022.

[37] Xueqin Zhang, Zhongqiang Luo, and Wenshi Xiao. Cnn-bilstm-dnn-based modulation recognition algorithm at low snr. Applied Sciences, 14(13):5879, 2024.

[38] ThienHuynhThe. Github - thienhuynhthe/mcnet: Mcnet: An efficient cnn architecture for robust automatic modulation classification.

[39] FatimaEzzahra Laghrissi, Samira Douzi, Khadija Douzi, and Badr Hssina. Intrusion detection systems using long short-term memory (lstm). Journal of Big Data, 8(1):65, 2021.

[40] Chuan Wang and Nianwen Xue. Getting the most out of amr parsing. In Proceedings of the 2017 conference on empirical methods in natural language processing, pages 1257–1268, 2017.

[41] Vairavan Srinivasan, Chikkannan Eswaran, Sriraam, and N. Artificial neural network based epileptic detection using time-domain and frequency-domain features. Journal of Medical Systems, 29:647–660, 2005.

[42] Mehrez Souden, Jacob Benesty, and Sofiene Affes. On optimal frequency-domain multichannel linear filtering for noise reduction. IEEE Transactions on audio, speech, and language processing, 18(2):260–276, 2009.

[43] Markus Endres, Asha Mannarapotta Venugopal, and Tung Son Tran. Synthetic data generation: A comparative study. In Proceedings of the 26th International Database Engineered Applications Symposium, pages 94–102, 2022.

[44] Alvaro Figueira and Bruno Vaz. Survey on synthetic data generation, evaluation methods and gans. Mathematics, 10(15):2733, 2022.

[45] Zhaoqing Pan, Weijie Yu, Xiaokai Yi, Asifullah Khan, Feng Yuan, and Yuhui Zheng. Recent progress on generative adversarial networks (gans): A survey. IEEE access, 7:36322–36333, 2019.

[46] Carl Doersch. Tutorial on variational autoencoders. arXiv preprint arXiv:1606.05908, 2016.

[47] Andoni Cortés, Clemente Rodríguez, Gorka Vélez, Javier Barandiarán, and Marcos Nieto. Analysis of classifier training on synthetic data for cross-domain datasets. IEEE Transactions on Intelligent Transportation Systems, 23(1):190–199, 2020.

[48] Grega Vrbančič and Vili Podgorelec. Transfer learning with adaptive fine-tuning. IEEE Access, 8:196197–196211, 2020.

[49] Nghi DQ Bui, Hung Le, Yue Wang, Junnan Li, Akhilesh Deepak Gotmare, and Steven CH Hoi. Codetf: One-stop transformer library for state-of-the-art code llm. arXiv preprint arXiv:2306.00029, 2023.

[50] Bingjie Liu, Qiancheng Zheng, Heng Wei, Jinxian Zhao, Haoyuan Yu, Yiyi Zhou, Fei Chao, and Rongrong Ji. Deep hybrid transformer network for robust modulation classification in wireless communications. Knowledge-Based Systems, 300:112191, 2024.

[51] Sepp Hochreiter and Jürgen Schmidhuber. Long short-term memory. Neural computation, 9(8):1735–1780, 1997.

[52] Anusha Lalitha, Shubhanshu Shekhar, Tara Javidi, and Farinaz Koushanfar. Fully decentralized federated learning. In Third workshop on bayesian deep learning (NeurIPS), volume 2, 2018.

[53] Yu Wang, Guan Gui, Haris Gacanin, Bamidele Adebisi, Hikmet Sari, and Fumiyuki Adachi. Federated learning for automatic modulation classification under class imbalance and varying noise condition. IEEE Transactions on Cognitive Communications and Networking, 8(1):86–96, 2021.